\documentclass{article}

\PassOptionsToPackage{numbers, compress}{natbib}

\usepackage[preprint]{neurips_2025}

\usepackage[utf8]{inputenc} 
\usepackage[T1]{fontenc}    
\usepackage{url}            
\usepackage{amsfonts}       
\usepackage{amsmath}        
\usepackage{amssymb}
\usepackage{nicefrac}       
\usepackage{microtype}      
\usepackage{graphicx}
\usepackage{bm}
\usepackage{amsthm}
\usepackage{makecell}
\usepackage{mathtools}
\usepackage{multirow}
\usepackage{wrapfig}
\usepackage{algorithm}
\usepackage{algpseudocode}
\usepackage[dvipsnames]{xcolor}
\usepackage{booktabs}       
\usepackage[colorlinks=true, citecolor=RedOrange, linkcolor=RedOrange]{hyperref}
\usepackage{cleveref}
\usepackage[super]{nth}

\newcommand{\dd}{{\rm d}}
\newcommand{\VVert}[1]{\left\Vert{#1}\right\Vert}
\newcommand{\AAngle}[1]{\left\langle{#1}\right\rangle}
\newcommand{\sss}[2]{{#1{\footnotesize{±#2}}}}

\newtheorem{theorem}{Theorem}

\newtheorem{lemma}{Lemma}

\newenvironment{autoproof}[1]
  {\begin{proof}[Proof of \autoref{#1}]}
  {\end{proof}}

\title{Conditioning Matters: 
 Training Diffusion Policies is Faster Than You Think}

\author{Zibin Dong\textsuperscript{$\heartsuit$}, Yicheng Liu\textsuperscript{$\clubsuit$}, Yinchuan Li\textsuperscript{$\diamondsuit$}, Hang Zhao\thanks{Corresponding authors: Hang Zhao (hangzhao@mail.tsinghua.edu.cn), Jianye Hao (jianye.hao@tju.edu.cn).}~~\textsuperscript{$\clubsuit$}, Jianye Hao\footnotemark[1]~~\textsuperscript{$\heartsuit,\diamondsuit$}\\
\textsuperscript{$\heartsuit$}Tianjin University, \textsuperscript{$\clubsuit$}Tsinghua University, \textsuperscript{$\diamondsuit$}Huawei Noah’s Ark Lab
}

\begin{document}

\maketitle

\begin{abstract}
Diffusion policies have emerged as a mainstream paradigm for building vision-language-action (VLA) models. Although they demonstrate strong robot control capabilities, their training efficiency remains suboptimal. In this work, we identify a fundamental challenge in conditional diffusion policy training: when generative conditions are hard to distinguish, the training objective degenerates into modeling the marginal action distribution, a phenomenon we term \textit{loss collapse}. To overcome this, we propose Cocos, a simple yet general solution that modifies the source distribution in the conditional flow matching to be condition-dependent. By anchoring the source distribution around semantics extracted from condition inputs, Cocos encourages stronger condition integration and prevents the loss collapse. We provide theoretical justification and extensive empirical results across simulation and real-world benchmarks. Our method achieves faster convergence and higher success rates than existing approaches, matching the performance of large-scale pre-trained VLAs using significantly fewer gradient steps and parameters. Cocos is lightweight, easy to implement, and compatible with diverse policy architectures, offering a general-purpose improvement to diffusion policy training.
\end{abstract}

\begin{figure}[h]
\vspace{-15pt}
    \centering
    \includegraphics[width=0.85\linewidth]{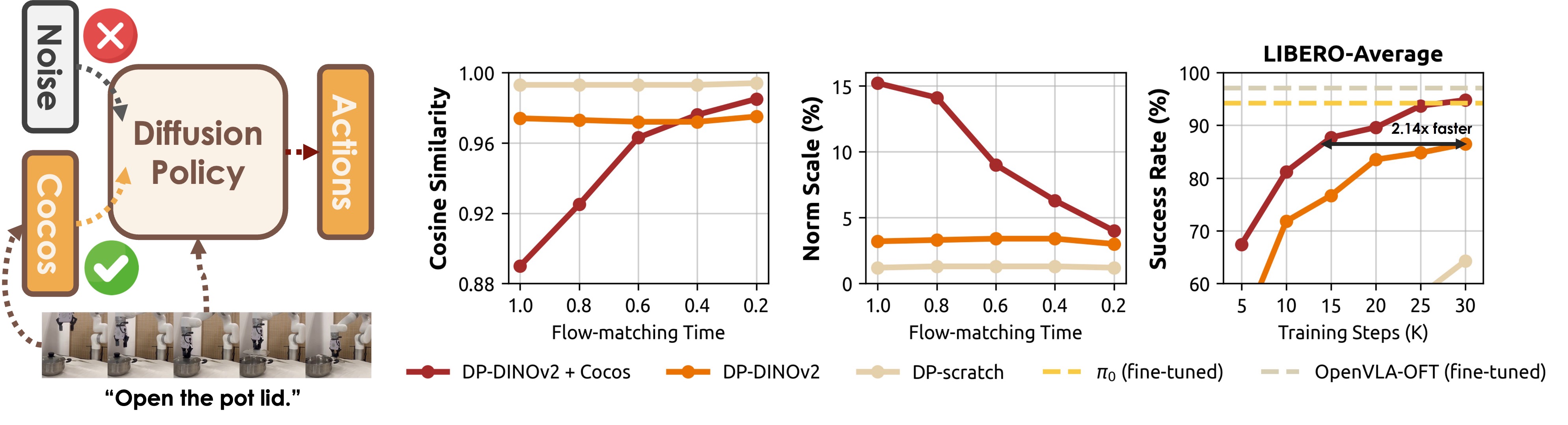}
    \vspace{-8pt}
    \caption{\small{\textbf{Fusing generative condition into the source distribution greatly simplifies diffusion policy training.} Diffusion policy trained with our method achieves $\pi_0$ performance on the LIBERO benchmarks with only 30K gradient steps, which is 2.14x faster than the vanilla model. We also show the cosine similarity and the norm scale change between the policy hidden states before and after injecting condition information, demonstrating that our method fundamentally compels the policy network to utilize condition information, rather than simply embedding conditions into the source distribution.}}
    \label{fig:motivation}
    \vspace{-15pt}
\end{figure}

\section{Introduction}
\vspace{-5pt}

Denoising generative models have emerged as a scalable approach for high-dimensional data generation \citep{song2021scorebased, ho2020ddpm, song2021ddim, lu2022dpmsolver, lu2023dpmsolverpp, lipman2023flow, tong2024improving, liu2023rectifiedflow}, significantly advancing Vision-Language-Action (VLA) models in robotics (VLAs\footnote{Unless otherwise specified, VLAs in this paper refer specifically to denoising generative VLA models.}) \citep{octo_2023, liu2025rdt, black2024pi0, chi2023diffusionpolicy, Ze2024DP3, li2025survey}. Mainstream VLA models frame robot control as a conditional generation problem: given vision-language inputs as conditions, the model generates appropriate robot action sequences. Despite action sequences being substantially lower-dimensional than typical generative content like images or text, training VLAs remains unexpectedly challenging and resource-intensive. Recent efforts to enhance VLA training efficiency have explored various directions, including leveraging more powerful vision-language encoders \citep{kim2024openvla, black2024pi0, qu2025spatialvla, kim2025openvlaoft, wen2024tinyvla, wen2024diffusionvla}, designing compact and expressive action tokenizers \citep{pertsch2025fast, wang2024omnijarvis}, and incorporating richer conditioning strategies \citep{huang2024rekep, ju2025roboabc, yuan2025robopoint, liu2025hybridvla, qi2025sofar}. These approaches converge on a critical insight: the core challenge may not lie in action generation itself, but rather in how models interpret and utilize conditional inputs.

To further investigate how condition inputs affect policy training, we empirically compare diffusion policies with visual encoders trained from scratch versus those initialized with DINOv2. We analyze the cosine similarity and norm scale change between policy network hidden states before and after condition information injection, correlating these metrics with LIBERO task success rates \Cref{fig:motivation}. Higher cosine similarity and lower norm scale change indicate weaker condition influence on the policy network, directly correlating with degraded performance (as observed comparing DP-DINOv2 versus DP-scratch). The results demonstrate a fundamental issue: policies facing difficult-to-interpret conditions \textit{actively omit conditional inputs}, producing actions disconnected from observations. This phenomenon persists even when using high-quality representations, though to a lesser degree.

While empirical results reveal this condition omission phenomenon, we conduct a theoretical analysis to identify its root cause in VLA training. We discover that when policy networks struggle to differentiate between generative conditions, the flow-matching objective degrades into an unconditional one that merely models the marginal action distribution. This loss collapse creates a destructive feedback loop: as the policy network begins ignoring conditional inputs, the training objective further degrades, reinforcing the network's tendency to discard conditions altogether rather than attempting to interpret them. To prevent this loss collapse, we introduce a novel conditional flow-matching approach with a \textbf{co}ndition-\textbf{co}nditioned \textbf{s}ource distribution (\textbf{Cocos}). Rather than adopting a standard Gaussian prior $q(z)$, Cocos anchors the source distribution around the semantics of each condition $q(z|c)$, theoretically preventing training loss collapse and forcing the policy network to remain responsive to condition inputs. As demonstrated in \Cref{fig:motivation}, diffusion policies trained with Cocos exhibit 2.14x faster training and substantially higher success rates. Critically, the lower cosine similarity and increased norm scale change indicate that Cocos fundamentally compels the policy network to utilize condition information, rather than simply embedding conditions into the source distribution.

Specifically, we implement Cocos by formulating the source distribution as a Gaussian with fixed standard deviation, where the mean derives from latent representations produced by a vision-language autoencoder (see \Cref{fig:cocos_main}). This approach introduces minimal architectural overhead, making it compatible with diverse VLA architectures and model scales. This simplicity, combined with strong theoretical guarantees and empirical benefits, positions Cocos as a general-purpose solution for preventing condition omission in VLA training.

To comprehensively validate Cocos, we conduct extensive evaluations across diverse settings: 70 simulation tasks from the LIBERO and MetaWorld benchmarks \citep{liu2023libero, yu2019metaworld}, 10 real-world tasks on the low-cost open-source SO-100 robot platform \citep{cadene2024lerobot}, and 10 tasks on the high-performance xArm robot platform. Our results demonstrate significant improvements in both manipulation success rates and learning efficiency. Through detailed case studies, we analyze loss collapse manifestations and the mechanisms through which Cocos enhances performance. Our contributions include:

\begin{figure}[t]
\vspace{-10pt}
    \centering
    \includegraphics[width=1.0\linewidth]{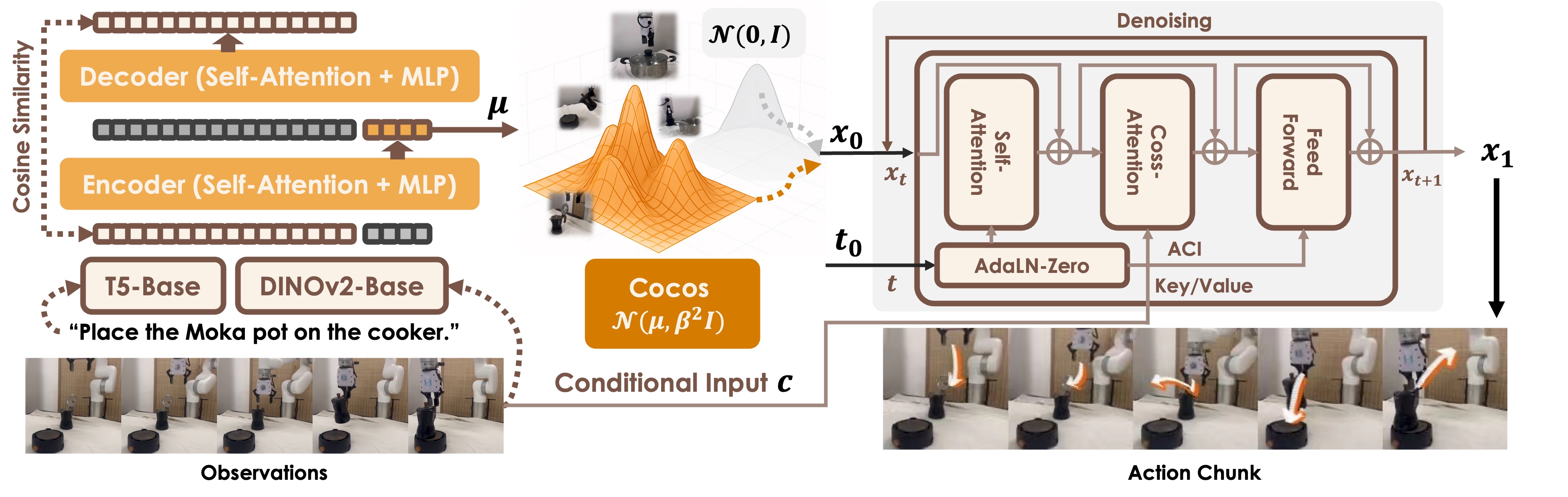}
    \vspace{-18pt}
    \caption{\small{\textbf{Diffusion Policy w/ Cocos.} Our approach requires only replacing the standard Gaussian source distribution with a condition-conditioned Gaussian. An autoencoder compresses condition representations to match dimensionality, providing the mean while maintaining a fixed standard deviation.}}
    \label{fig:cocos_main}
    \vspace{-15pt}
\end{figure}

\begin{itemize}
    \vspace{-5pt}
    \item We formulate the mathematical framework of flow matching with generative conditions and demonstrate that policy networks omit conditions when they are difficult to distinguish.
    \vspace{-2pt}
    \item We introduce \textbf{Cocos}, a simple yet effective source distribution modification that prevents loss collapse and significantly improves diffusion policy training efficiency and performance.
    \vspace{-5pt}
    \item We establish a comprehensive evaluation benchmark across diverse settings, including simulation tasks from LIBERO and MetaWorld, real-world tasks on a low-cost open-source robot (SO100), and tasks on a high-performance robot (xArm). Our results demonstrate consistent performance improvements across these varied platforms, validating Cocos as a general-purpose, plug-and-play solution for enhancing diffusion policy training.
    \vspace{-5pt}
\end{itemize}

\section{Preliminaries}

\textbf{Problem Formulation.} Our goal is to train a VLA $\pi_\theta(a_{1:H}|\mathcal E(o))$ that maps an observation $ o$ to sequences of future robot actions $a_{1:H}$ by imitating expert demonstrations. Here, $o$ typically includes current and historical images as well as language instructions, while $\mathcal E$ represents a pre-trained vision-language encoder. In this paper, we focus on denoising generative model VLAs.

\textbf{Conditional Flow Matching.} Assume a smooth time-varying vector field $u:[0,1]\times\mathbb R^d\rightarrow \mathbb R^d$ defines an ordinary differential equation (ODE) $\dd x=u_t( x)\dd t$, which pushforward samples from source distribution $p_0$ to the target distribution $p_1$. The density transported along $u$ from time $0$ to $t$ is denoted as $p_t:[0,1]\times\mathbb R^d\rightarrow\mathbb R$. $p_t$ and $u_t$ satisfy the \textit{continuity equation}: $\partial p/\partial t=-\nabla(p_t\cdot u_t)$. Suppose that the marginal probability path $p_t(x)$ is a mixture of probability paths $p_t( x| z)$, that is $p_t( x)=\int p_t( x| z)q( z)\dd  z$, where $q( z)$ is some distribution over the conditioning variable. If $p_t( x| z)$ is generated by the vector field $u_t( x| z)$ from initial conditions $p_0( x| z)$, then the vector field
\begin{equation}
    u_t( x):=\mathbb E_{q( z)}\frac{u_t( x| z)p_t( x| z)}{p_t( x)}
\end{equation}
generates the probability path $p_t( x)$. Let $v_\theta:[0,1]\times\mathbb R^d\rightarrow\mathbb R^d$ be a time-dependent vector field parameterized as a neural network $\theta$. Define the conditional flow matching (CFM) objective:
\begin{equation}
    \mathcal L_{\text{CFM}}(\theta):=\mathbb E_{t,q(z),p_t(x|z)}\left\Vert v_\theta(t, x)-u_t(x|z)\right\Vert^2.
\end{equation}
In VLA models, a typical choice sets $q(z)=q(x_0)q(x_1)$, corresponding to an independent coupling of the source and target distributions, and the conditional flow is defined by
\begin{equation}
    p_t(x|z)=p_t(x|x_1,x_0),~u_t(x|z)=u_t(x|x_1,x_0),
\end{equation}
where the source distribution is a standard Gaussian. By optimizing CFM objective, $v_\theta(t, x)$ can converge to the vector field $u_t(x)$, with which we can solve the ODE and sample action chunks. \textit{ Note that this is a general formulation covering a wide range of denoising generative models, e.g., Var. Exploding \citep{song2020vesde}, Var. Preserving \citep{ho2020ddpm} diffusion models, flow matching models \citep{lipman2023flow}, rectified flow models \citep{liu2023rectifiedflow}, etc. So our following discussion can be easily transferred to various policy backbones.}

\section{How Generative Conditions Can Lead to Loss Collapse}

While conditional flow matching offers a general framework for transporting probability distributions, it does not inherently consider condition inputs. Prior works adopt a straightforward strategy: injecting the generative condition directly into the policy network \citep{chisari2024fmpolicy1, zhang2025fmpolicy2, zhang2024fmpolicy3}. However, the implications of this approach have received little theoretical attention. In this section, we first analyze how the underlying flow formulation changes when generative conditions are introduced, then demonstrate that this seemingly benign modification can lead to a critical failure mode: loss collapse, where the training objective degenerates and the model learns to omit the generative condition altogether.

\subsection{Flow with Generative Conditions}
We consider a conditional vector field $u_t(x|c)$ that transports the source distribution $p_0(x|c)$ to the target $p_1(x|c)$, where $c$ is the generative condition. Let $p_t(x|z,c)$ denote the time-varying intermediate density under generation condition $c$ and flow condition $z$, with its associated vector field $u_t(x|z,c)$. Then, the overall conditional velocity can be expressed as:
\begin{equation}
    u_t(x|c):=\mathbb E_{q(z)}\frac{u_t(x|z,c)p_t( x|z,c)}{p_t(x|c)}.
\end{equation}
Suppose we introduce $c$ into the neural estimator $v_\theta(t,x,c)$ of the conditional velocity $u_t(x|c)$. The corresponding training objective becomes:
\begin{equation}\label{eq:CFMc_general}
    \mathcal L_{\text{CFMc}}(\theta):=\mathbb E_{t,q(z),p_t(x|z,c)}\left\Vert v_\theta(t, x, c)-u_t(x|z, c)\right\Vert^2.
\end{equation}
\begin{lemma}\label{thm:1}
If $p_t(x|c)>0$ for all $x\in\mathbb R^d$ and $t\in[0,1]$, up to a constant independent of $\theta$, objective $\mathbb E_{t,p_t(x|c)}\left\Vert v_\theta(t, x, c)-u_t(x|c)\right\Vert^2$ and $\mathbb E_{t,q(z),p_t(x|z,c)}\left\Vert v_\theta(t, x, c)-u_t(x|z, c)\right\Vert^2$ are equal.
\end{lemma}

By optimizing \Cref{eq:CFMc_general}, we can use $v_\theta(t,x,c)$ to estimate $u_t(x|c)$ and solve the ODE for generation. In practice, a common formulation of $q(z)$ in \Cref{eq:CFMc_general} is $q(z)=q(x_0)q(x_1,c),~z=(x_1, x_0,c)$. This setup corresponds to sampling a condition-action pair from the dataset and perturbing the action with Gaussian noise during training. Under this formulation, the conditional flow and velocity field are defined by:
\begin{equation}
    p_t(x|z,c)=\left\{\begin{array}{cc}
        p_t(x|x_1,x_0)&,~c\in z \\
        0&,~c\notin z 
    \end{array}\right.~,~\text{and} ~
    u_t(x|z,c)=\left\{\begin{array}{cc}
        u_t(x|x_1,x_0)&,~c\in z \\
        0&,~c\notin z 
    \end{array}\right.
\end{equation}
This reduces the objective to that implemented in previous works \citep{chisari2024fmpolicy1, zhang2025fmpolicy2, zhang2024fmpolicy3}:
\begin{equation}\label{eq:CFMc}
    \mathcal L_{\text{CFMc}}(\theta):=\mathbb E_{t,q(x_0),q(x_1,c),p_t(x|x_1,x_0)}\left\Vert v_\theta(t, x, c)-u_t(x|x_1,x_0)\right\Vert^2.
\end{equation}

\begin{figure}
    \centering
    \includegraphics[width=0.9\linewidth]{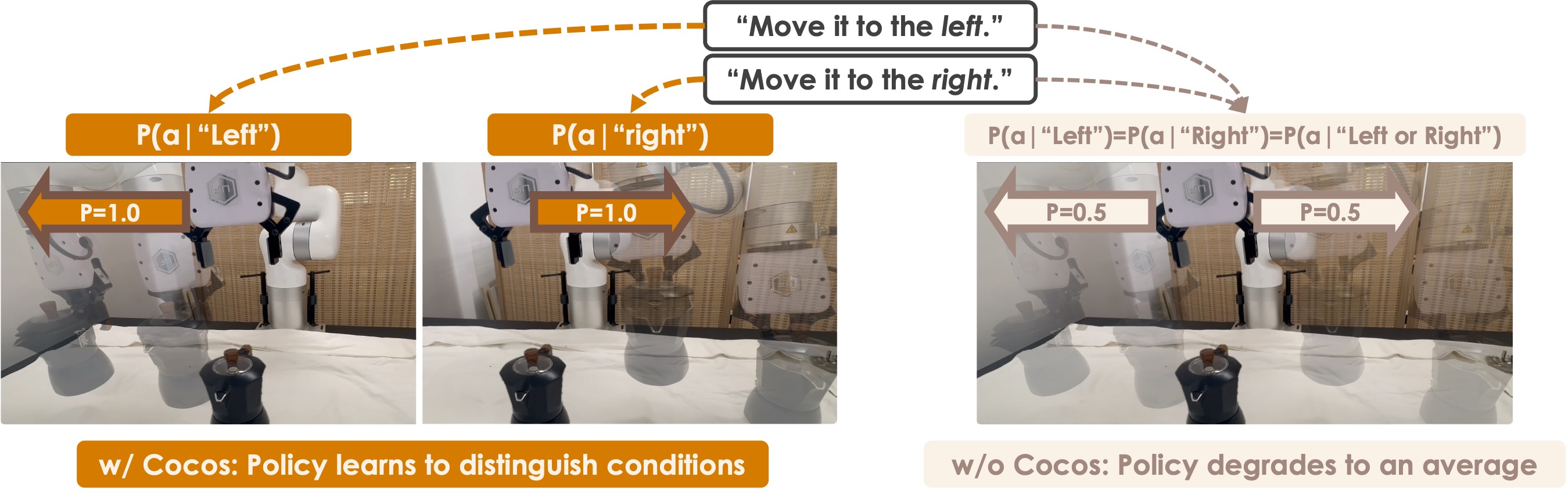}
    \caption{\small{\textbf{Loss collapse causes the policy to degrade to an average.} The policy omits language instructions (`Move it to the left' or `Move it to the right') and yields actions based on their frequency in the training data. Cocos prevents loss collapse, enabling the policy to produce distinct actions corresponding to `Left' and `Right'.}}
    \label{fig:loss_collaspe}
    \vspace{-10pt}
\end{figure}

\subsection{Loss Collapse}

While this conditional objective encourages the model to learn condition-dependent vector fields, a critical failure mode can occur: if the model fails to distinguish the condition information in $\mathcal C$ (e.g., for any $c_1, c_2\in\mathcal C$,  \(\|v_\theta(t,x,c_1)-v_\theta(t,x,c_2)\|\le\epsilon\)), the objective effectively collapses into a marginal form on $\mathcal C$ that no longer depends on $c$. 
\begin{theorem}[Gradient Contraction under Independent Sampling]\label{thm:2}
Under the independent sampling measure
\[
\mu(\dd y)=q(x_1,c)\,q(x_0)\,p_t(x| x_1,x_0)\,\dd x_0\dd x_1\dd x,
\]
the difference of the score gradients for any two conditions \(c_1,c_2\in\mathcal C\) admits the bound
\[
\bigl\|\nabla_\theta\mathcal L_{\mathrm{CFMc}}(\theta,c_1)
 -\nabla_\theta\mathcal L_{\mathrm{CFMc}}(\theta,c_2)\bigr\|
 \;\le\;2\,(M+K D)\,\epsilon,
\]
provided that \(\|\nabla_\theta v_\theta\|\le M\), \(\|d\|\le D\), and the model output satisfies \(\|v_\theta(t,x,c_1)-v_\theta(t,x,c_2)\|\le\epsilon\) with its gradient being output‐sensitive Lipschitz of constant \(K\). The velocity estimator $v_\theta(t,x,c)$ tends to converge to a $c$-independent function $v^*(t,x)$ that minimizes the averaged objective:
\[
    v^*(t,x) := \arg\min_{v}~\mathbb{E}_{c\in\mathcal C} \mathbb{E}_{z\sim\mu} \left[\|v(t,x) - u_t(x|x_1,x_0)\|^2\right].
\]

By contrast, when samples are drawn from the conditional measure
\[
\mu_c(\dd y)=q(x_1,c)\,q(x_0| c)\,p_t(x| x_1,x_0)\,\dd x_0 \dd x_1 \dd x,
\]
the same gradient difference can become arbitrarily large unless additional coupling assumptions on \(q(x_0| c)\) are imposed.
\end{theorem}

This result reveals a vicious cycle: as the policy network becomes confused by condition inputs, the objective degenerates into an unconditional one. This further reinforces the network's tendency to discard difficult-to-interpret conditions, leading to models that appear to be well-trained but fail to comprehend conditional cues during deployment. As shown in \Cref{fig:loss_collaspe}, loss collapse causes the policy to average actions. In the next section, we introduce Cocos, a simple method to prevent it.

\vspace{-5pt}
\section{Method}
\vspace{-5pt}
\label{sec:method}

In \Cref{thm:2}, we identify that loss collapse arises when $c$ becomes indistinguishable to the policy network. This collapse is exacerbated by the standard choice of sampling $z$ from $q(z) = q(x_1, c)q(x_0)$, which decouples $x_0$ from the condition. If $v_\theta(t, x, c) \approx v_\theta(t, x)$, the expectation in the objective effectively marginalizes over $c$, causing it to degenerate into an unconditional one.

To prevent it, we propose a simple yet effective modification: using a \textbf{co}ndition-\textbf{co}nditioned \textbf{s}ource distribution (\textbf{Cocos}) by sampling $z$ from $q(z)=q(x_1,c)q(x_0|c)$. As shown in \Cref{thm:2}, this change directly avoids the loss collapse and enforces condition-awareness during training. While $q(x_0 | c)$ can be designed in various ways, in this work, we choose to maintain the overall framework's simplicity by setting it as a Gaussian with fixed standard deviation, with the following objective:
\begin{equation}\label{eq:cocos}
\mathcal L_{\text{Cocos}}(\theta):=\mathbb E_{t,\textcolor{RedOrange}{q(x_0|c)},q(x_1,c),p_t}\left\Vert v_\theta-u_t\right\Vert^2,~\text{where}~q(x_0|c)=\mathcal N(x_0;\alpha F_\phi(\mathcal E(c)), \beta^2I),
\end{equation}
where $F_\phi$ is a feature encoder that maps the condition representations to the action space dimensionality. The scalar $\alpha$ controls the strength of the condition prior, and $\beta$ adjusts the uncertainty. Setting $\alpha=0,\beta=1$ recovers the commonly used formulation with a standard Gaussian prior. Incorporating Cocos requires only replacing the standard Gaussian source distribution with a condition-conditioned one. \Cref{alg:code} clearly shows the differences in training and inference pipelines.

In practice, we adopt an autoencoding objective on condition embeddings to train $F_\phi$. Specifically, we minimize the negative cosine similarity between the original and reconstructed embeddings:
\begin{equation}
\mathcal L(\phi)=-\mathbb E_{c}\text{Sim}\left(G_\phi(F_\phi(\mathcal E(c))),\mathcal E(c)\right),
\end{equation}
where $F_\phi$ and $G_\phi$ are the encoder and decoder networks, respectively. As a default setting, we adopt a two-stage training process: we first train the source distribution (i.e., the autoencoder) and then fix it during policy training. However, in practical scenarios, this two-step pipeline may introduce additional inflexibility. To address this, we also explore a joint training alternative. Simultaneous training of this autoencoder and the policy can be unstable due to the evolving distribution of $x_0$. To mitigate this, we adopt an Exponential Moving Average (EMA) strategy. We maintain a target network $F_{\phi^-}$, updated via EMA from $F_\phi$, and use this EMA copy during policy training, ensuring stability. In practice, this approach achieves performance comparable to the two-stage strategy, which we will discuss in detail in the experimental section.

We implement our policy network using a compact Robot Diffusion Transformer (RDT) \citep{liu2025rdt}. The diffusion process follows a linear interpolation schedule \citep{lipman2023flow}:
\begin{equation}
\label{eq:diffusion_implement}
    p_t(x|x_1,x_0)=\mathcal N(x;tx_1+(1-t)x_0,\sigma^2),~~u_t(x|x_1,x_0)=x_1-x_0.
\end{equation}
We select these design choices as they have been widely adopted in advanced VLA models \citep{liu2025rdt, black2024pi0}.

\begin{algorithm}[h]
\caption{\small Training and Inference Pseudocode for Diffusion Policy with Cocos}\label{alg:code}
    \begin{algorithmic}[1]
    \footnotesize
    \Require Policy Network $v_\theta$, Condition Network $\mathcal E$, Pre-trained condition encoder $F_\phi$
    \While {not coverge} \textcolor{gray}{\texttt{\# Training}}
        \State $(x_1, c)\sim q(x_1,c), ~t\sim \mathcal U(0,1),~$\textcolor{RedOrange}{$x_0\sim q(x_0|c)$~~\Cref{eq:cocos}} \textcolor{gray}{\# $x_0\sim \mathcal N(0, I)$~~\texttt{(without Cocos)}}
        \State $\theta \leftarrow \theta +\nabla_\theta \mathcal{L_{\text{CFMc}}}(\theta)$~~\Cref{eq:CFMc}
    \EndWhile
    \While {not done} \textcolor{gray}{\texttt{\# Inference}}
        \State Observe $c,~$\textcolor{RedOrange}{$x_0\sim q(x_0|c)$~~\Cref{eq:cocos}} \textcolor{gray}{\# $x_0\sim \mathcal N(0, I)$~~\texttt{(without Cocos)}}
        \State Solve ODE $\dd x_t=v_\theta(t,x_t,c)\dd t$ from $t=0$ to $t=1$.
        \State $a\leftarrow x_1$; Execute action $a$
    \EndWhile
    \end{algorithmic}
\end{algorithm}

\vspace{-5pt}
\section{Experiments}
\vspace{-5pt}

We conduct comprehensive experiments to evaluate the effectiveness of Cocos in both simulated and real-world robot manipulation tasks. The experiments are designed to answer three core questions:

\textbf{(RQ1)} Does Cocos improve diffusion policies' training efficiency and final performance?

\textbf{(RQ2)} Does it mitigate the loss collapse phenomenon?

\textbf{(RQ3)} Does it facilitate policy learning in real-world settings across heterogeneous robot platforms?

\subsection{Experimental Setup} 

\textbf{Diffusion Policy.} Our diffusion policy (DP) adopts a compact RDT policy network of approximately 40M parameters. The vision-language condition inputs are encoded using a DINOv2-Base \citep{oquab2024dinov} and a T5-Base \citep{2020t5}. We denote the resulting model as DP-DINOv2. To evaluate the effectiveness of Cocos, we compare three key variants: (1) DP-scratch, which is trained without any pre-trained vision encoder; (2) DP-DINOv2, which incorporates frozen DINOv2 features but uses the standard source distribution; and (3) DP-DINOv2 with Cocos, our full method with a condition-conditioned source distribution. The autoencoder used to define the source distribution in Cocos is implemented using a single-layer Transformer for both encoder and decoder components. 

\textbf{Benchmarks.} Our simulation evaluations are based on the LIBERO and MetaWorld benchmarks. LIBERO includes 40 tasks in four task suites: \textit{Goal}, \textit{Spatial}, \textit{Object}, and \textit{Long}, to test different policy generalizations. MetaWorld includes 30 tasks from various difficulty levels. For real-world experiments, we deploy the models on two robot platforms: The SO100 robot (low-cost, open-sourced, equipped with dual RGB cameras) evaluated on 10 tasks in four suites: \textit{Pick\&Place}, \textit{MoveTo}, \textit{Wipe}, and \textit{Unfold}; The xArm robot (higher-precision, equipped with one Intel RealSense L515 LiDAR camera) evaluated on 10 tasks in suites: \textit{Pick\&Place}, \textit{Pot}, \textit{Pour}, and \textit{Moka}. We show detailed task configurations in \Cref{appendix:benchmarks}.

\subsection{Overall Comparison (RQ1)}

\begin{figure}
    \centering
    \makebox[\textwidth][c]{
    \includegraphics[width=1.0\linewidth]{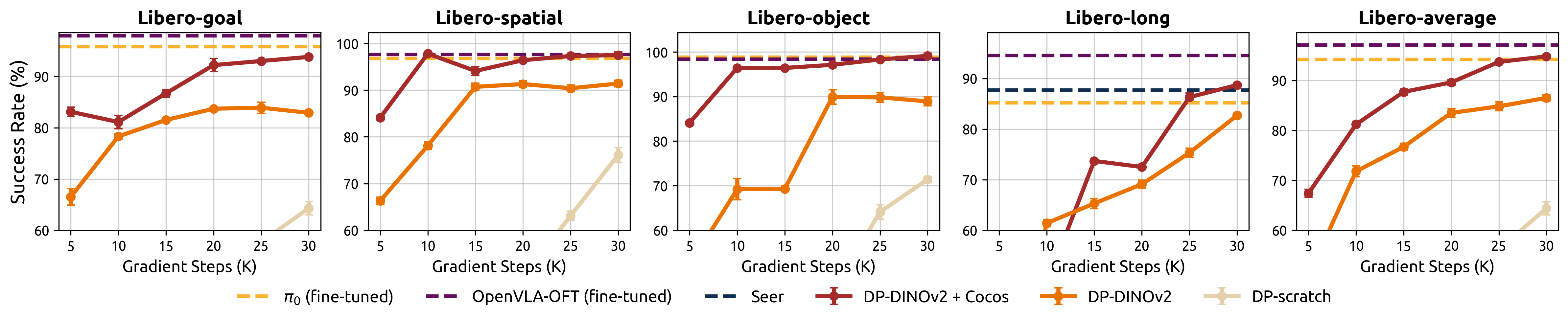}}
    \vspace{-15pt}
    \caption{\small{\textbf{Learning curve on LIBERO benchmark.} Dashed line scores are reported by \citet{kim2025openvlaoft}.}}
    \label{fig:libero_main}
\end{figure}

\begin{table}[t]
\caption{\small{\textbf{Success rate on LIBERO benchmark.} Each of the four task suites includes 10 tasks, and we evaluate 150 trials for each task. The \nth{1}, \nth{2} highest scores are emphasized with bold and the \nth{3} with underline. Cocos (0.2 std) is our default configuration, and other Cocos variants serve for an ablation study.}}
\label{tab:libero_results}
\scalebox{0.7}{
\begin{tabular}{c|lccccccc}
\toprule
\multicolumn{1}{l}{}    &         & DP-scratch & DP-DINOv2 & \underline{Cocos (0.2 std)} & Cocos (0.1 std) & Cocos (0.4 std) & Cocos (VAE)   & Cocos (EMA)   \\
\midrule
\multirow{4}{*}{LIBERO} & Goal    
& \sss{64.3}{1.3} & \sss{82.9}{0.4} & \textbf{\sss{93.8}{0.3}} & \sss{50.9}{1.8} & \underline{\sss{92.3}{0.9}} & \textbf{\sss{92.5}{0.3}} & \sss{90.5}{0.4} \\
& Spatial & \sss{76.1}{1.6} & \sss{91.4}{0.7} & \textbf{\sss{97.5}{0.7}} & \underline{\sss{97.1}{0.4}} & \sss{96.7}{0.5} & \sss{96.1}{0.7} & \textbf{\sss{97.5}{0.2}} \\
& Object  & \sss{71.4}{0.7} & \sss{88.9}{1.0} & \textbf{\sss{99.1}{0.6}} & \sss{79.3}{0.8} & \underline{\sss{97.6}{1.0}} & \textbf{\sss{98.1}{0.7}} & \sss{96.9}{0.5} \\
& Long & \sss{45.9}{1.6} & \sss{82.7}{0.2} & \textbf{\sss{88.7}{0.1}} & \sss{82.7}{0.5} & \textbf{\sss{89.5}{0.4}} & \underline{\sss{88.2}{0.6}} & \sss{87.3}{0.6} \\
\midrule
\multicolumn{2}{c}{Average}       & 64.4       & 86.5      & \textbf{94.8}               & 77.5            & \textbf{94.0}   & \underline{93.8}    & 93.0     \\  
\bottomrule
\end{tabular}}
\vspace{-10pt}
\end{table}

\begin{table}[t]
\caption{\small{\textbf{Success rate on MetaWorld benchmark.} We evaluate 150 trials for each task. The highest scores are emphasized in bold. We take a multi-task setting, using language instructions to differentiate each task.}}
\label{tab:metaworld_results}
\centering

\makebox[\textwidth][c]{

\scalebox{0.51}{

\begin{tabular}{l|cccccccccc|c}

\toprule

          & button-press   & button-press-td & button-press-td-w & button-press-w & coffee-button  & door-open     & door-lock     & door-unlock      & drawer-close   & drawer-open    & \textbf{Average}  \\

\midrule

w/ Cocos  & \textbf{\sss{100.0}{0.0}} & \sss{95.3}{0.9}                 & \textbf{\sss{94.7}{3.4}}             & \sss{97.3}{2.5}              & \textbf{\sss{100.0}{0.0}} & \textbf{\sss{99.3}{0.9}} & \textbf{\sss{44.7}{1.9}} & \textbf{\sss{93.3}{4.1}}    & \textbf{\sss{100.0}{0.0}} & \textbf{\sss{100.0}{0.0}} & \textbf{\sss{74.8}{0.9}}     \\

w/o Cocos & \sss{97.3}{2.5}           & \textbf{\sss{96.0}{1.6}}        & \sss{91.3}{4.1}                      & \textbf{\sss{99.3}{0.9}}     & \sss{97.3}{2.5}           & \sss{77.3}{5.7}          & \sss{3.3}{0.9}           & \sss{67.3}{5.2}             & \textbf{\sss{100.0}{0.0}} & \textbf{\sss{100.0}{0.0}} & \sss{59.5}{3.8}              \\

scratch   & \sss{48.7}{8.1}           & \sss{41.3}{6.6}                 & \sss{32.7}{6.6}                      & \sss{82.0}{5.9}              & \sss{53.3}{0.9}           & \sss{50.7}{4.7}          & \sss{23.3}{2.5}          & \sss{24.7}{3.4}             & \sss{88.7}{0.9}           & \sss{57.3}{5.0}           & \sss{33.6}{3.4}              \\

\midrule

          & faucet-close   & faucet-open          & handle-press              & handle-press-s & window-close   & window-open   & handle-pull   & handle-pull-s & basketball     & bin-picking    & \multirow{8}{*}{} \\

\midrule

w/ Cocos  & \textbf{\sss{82.0}{2.8}} & \textbf{\sss{100.0}{0.0}} & \textbf{\sss{99.3}{0.9}} & \textbf{\sss{100.0}{0.0}} & \textbf{\sss{100.0}{0.0}} & \textbf{\sss{94.0}{3.3}} & \textbf{\sss{2.7}{0.9}} & \textbf{\sss{9.3}{4.7}} & \textbf{\sss{80.7}{8.6}} & \textbf{\sss{64.0}{6.2}} &                   \\

w/o Cocos & \sss{80.7}{3.8}           & \sss{84.0}{4.3}                 & \sss{98.0}{1.6}                      & \sss{98.0}{1.6}              & \textbf{\sss{100.0}{0.0}} & \sss{89.3}{4.1}          & \sss{1.3}{0.9}           & \sss{0.0}{0.0}              & \sss{0.7}{0.9}            & \sss{46.0}{5.9}           &                   \\

scratch   & \sss{36.7}{3.4}           & \sss{79.3}{1.9}                 & \sss{76.0}{2.8}                      & \sss{53.3}{3.4}              & \sss{60.7}{1.9}           & \sss{26.0}{0.0}          & \sss{2.0}{1.6}           & \sss{0.0}{0.0}              & \sss{0.0}{0.0}            & \sss{10.0}{4.3}           &                   \\

\midrule

          & coffee-pull    & coffee-push          & dial-turn                 & hammer            & sweep          & sweep-into    & soccer        & shelf-place      & disassemble    & stick-push     &                   \\

\midrule

w/ Cocos  & \textbf{\sss{79.3}{6.2}} & \textbf{\sss{60.7}{5.0}} & \textbf{\sss{60.0}{2.8}} & \textbf{\sss{76.7}{6.6}} & \textbf{\sss{72.0}{8.5}} & \textbf{\sss{66.0}{8.6}} & \textbf{\sss{14.0}{3.3}} & \textbf{\sss{26.7}{0.9}} & \sss{39.3}{1.9} & \textbf{\sss{94.0}{4.3}} &                   \\

w/o Cocos & \sss{72.7}{4.7}           & \sss{55.3}{5.2}                 & \sss{24.0}{4.3}                      & \sss{72.7}{1.9}              & \sss{9.3}{2.5}            & \sss{20.0}{3.4}          & \textbf{\sss{14.0}{0.9}} & \sss{3.3}{2.5}              & \textbf{\sss{72.7}{4.9}}  & \sss{12.7}{1.9}           &                   \\

scratch   & \sss{21.3}{6.8}           & \sss{19.3}{5.7}                 & \sss{2.7}{0.9}                       & \sss{25.3}{2.5}              & \sss{10.0}{4.9}           & \sss{18.0}{3.3}          & \sss{6.7}{1.9}           & \sss{2.7}{0.9}              & \sss{14.7}{2.5}           & \sss{40.0}{2.8}           &        \\

\bottomrule

\end{tabular}}}

\end{table}

We evaluate the effectiveness of Cocos on two widely-used simulation benchmarks: LIBERO and MetaWorld. For LIBERO, we also include comparisons with fine-tuned $\pi_0$ \citep{black2024pi0}, OpenVLA-OFT \citep{kim2025openvlaoft}, and Seer \citep{tian2025seer}. $\pi_0$ and OpenVLA-OFT are SOTA large-scale pre-trained VLA models, featuring flow-matching and next-token prediction, respectively. Seer is not an end-to-end VLA model but predicts future frames to infer actions. We report the results in \Cref{tab:libero_results}, \Cref{tab:metaworld_results} and \Cref{fig:libero_main}.

\textcolor{RedOrange}{\textbf{\textit{Finding 1:}}} \textbf{Cocos significantly improves training efficiency and policy performance.} Across all LIBERO suites, Diffusion policy (DP) trained with Cocos outperforms its counterparts without Cocos. Notably, Cocos-enabled models reach high success rates (e.g., >95\% on \textit{Spatial} and \textit{Object} suites) within only 5–10K training steps, achieving convergence roughly \textbf{2.14x} faster than DP-DINOv2. On average, Cocos improves LIBERO success rates by \textbf{8.3\%}. Across all MetaWorld tasks, it also outperforms the counterpart by \textbf{25.7\%} gain, showing consistent performance gains.

\textcolor{RedOrange}{\textbf{\textit{Finding 2:}}} \textbf{Cocos enables compact models to rival large-scale pre-trained VLAs.} Despite using fewer parameters and fewer gradient steps, DP with Cocos performs comparably to large-scale models like $\pi_0$ and OpenVLA-OFT. As described in \citet{kim2025openvlaoft}, while OpenVLA-OFT takes 50-150K gradient steps for fine-tuning and $\pi_0$ takes 100-250K steps, our model reaches similar performance within 30K steps. On the challenging \textit{Long} suite, Cocos even outperforms Seer, a non-end-to-end model with a larger parameter count. These results demonstrate that proper condition integration, rather than model scale alone, can be a major driver of performance in VLA training.

\textcolor{RedOrange}{\textbf{\textit{Finding 3:}}} \textbf{Cocos encourages deeper condition utilization rather than just injecting priors.} To understand the nature of the improvement, we examine internal representations of the policy network before and after condition injection (\Cref{fig:motivation}). Specifically, we measure cosine similarity and norm scale changes of hidden states (see \Cref{appendix:metrics} for details). A lower similarity and a higher norm shift reflect greater responsiveness to conditioning inputs. As shown in \Cref{fig:motivation}, these metrics are consistently aligned with higher task success rates. This suggests that the benefit of Cocos does not stem merely from injecting condition priors into the source distribution. Instead, it actively compels the policy network to differentiate between and respond to condition inputs. If the gains were due only to initialization, we would expect weaker condition dependence, contrary to what we observe.

\begin{figure}[t]
    \centering
    \includegraphics[width=1.0\linewidth]{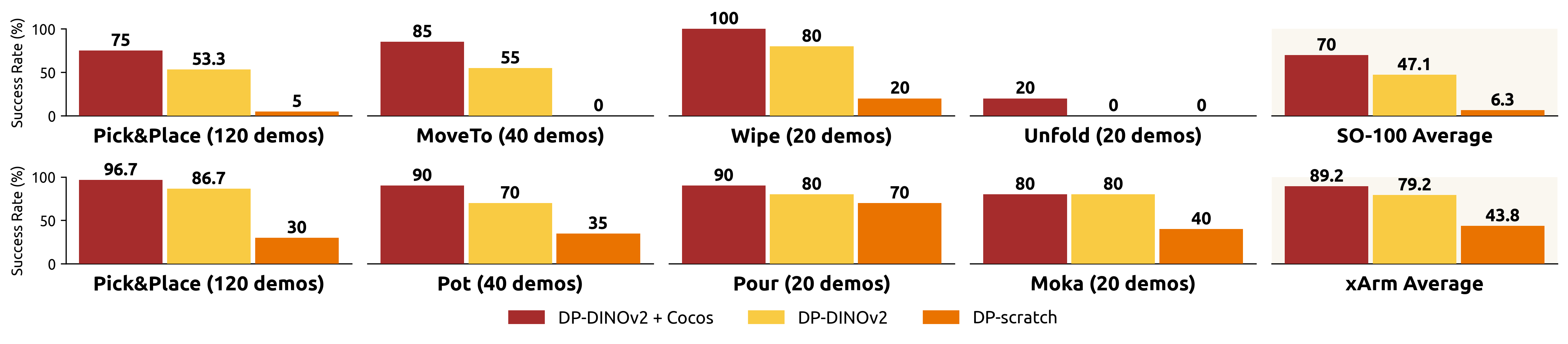}
    \vspace{-15pt}
    \caption{\small{\textbf{Evaluation results on SO100 and xArm platforms.} We collect 4 task suites or 10 tasks for each robot platform. Each task provides 20 demonstrations and is tested over 10 trials.}}
    \label{fig:cocos_real_robot_score}
    \vspace{-15pt}
\end{figure}

\begin{figure}[t]
    \centering
    \makebox[\textwidth][c]{
    \includegraphics[width=1.0\linewidth]{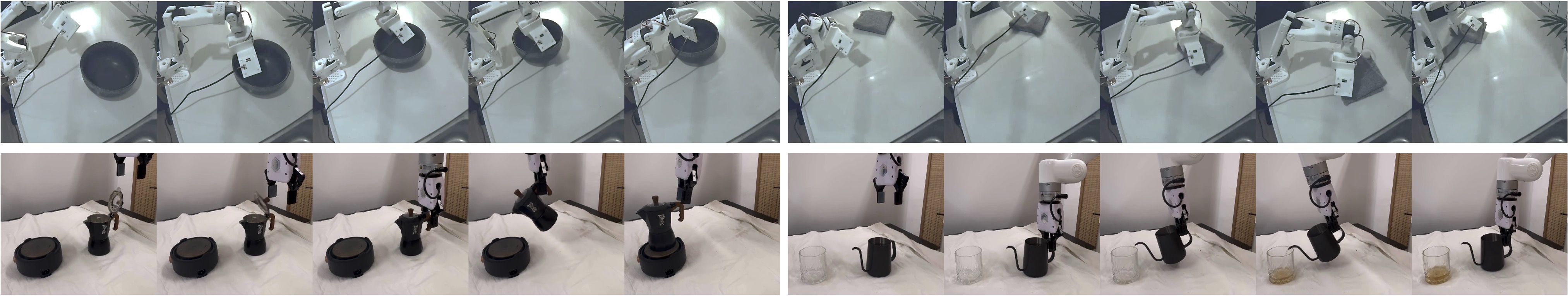}}
    \vspace{-15pt}
    \caption{\small{\textbf{Rollout samples of DP with Cocos on real-robot manipulation,} including \textit{MoveTo (Bowl)} (Top Left), \textit{Wipe} (Top Right), \textit{Moka} (Bottom Left), and \textit{Pour} (Bottom Right).}}
    \label{fig:real_robot_demo}
    \vspace{-5pt}
\end{figure}

\subsection{Real-World Experiments (RQ2)}

To assess the practicality of our approach, we deploy diffusion policies trained with and without Cocos on two real-world robotic platforms: \textbf{SO100}, a low-cost and open-source arm, and \textbf{xArm}, a widely used high-precision industrial robot. The two platforms offer complementary insights: SO100 lowers the entry barrier for research and reproducibility, while xArm reflects deployment scenarios with more demanding task complexity and control fidelity.

As shown in \Cref{fig:cocos_real_robot_score} and \Cref{fig:real_robot_demo}, Cocos consistently improves task success rates across all task suites on both robots. The improvement is particularly pronounced on SO100, where the relatively lightweight structure makes it more sensitive to inaccuracies in action generation. We observe that both policy variants, with and without Cocos, are capable of generating smooth and coherent trajectories. However, the main failure modes of the baseline model are closely linked to misinterpretation of condition inputs, for example, confusing spatial references in language, or failing to correctly localize an object in the visual scene. In contrast, these types of errors are reduced when using Cocos, indicating improved condition understanding and robustness. These trends align with our theoretical motivation: condition misalignment during training can lead to loss collapse and unreliable behavior at inference time. The fact that Cocos provides consistent improvements across two distinct robotic platforms further highlights its practical value. In the next section, we present detailed case studies to examine these qualitative differences in greater depth.

\subsection{Case Studies (RQ3)}

To better understand the mechanisms behind Cocos's effectiveness, we conduct qualitative case studies on both simulation and real-world tasks, focusing on common failure patterns and condition sensitivity.

\begin{wrapfigure}{t}{0.5\textwidth}
\centering
    \includegraphics[width=0.5\textwidth]{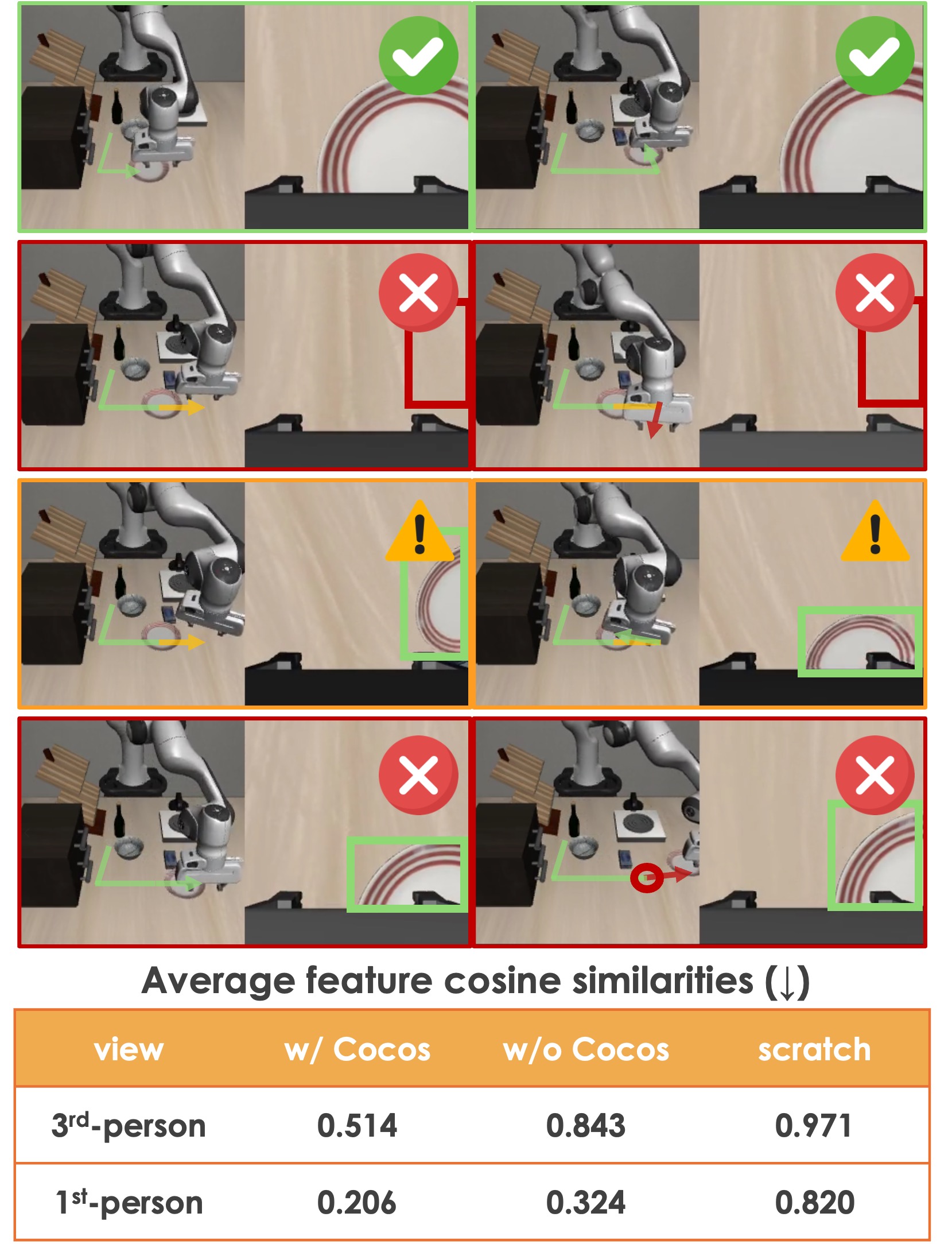}
    \vspace{-15pt}
    \caption{\small{\textbf{Case study of LIBERO plate movement.} }}
    \label{fig:libero_case_study}
    \vspace{-10pt}
\end{wrapfigure}

\textcolor{RedOrange}{\textbf{\textit{Finding 1:}}} \textbf{Cocos improves utilization of 3rd-person visual inputs in simulation.} In the LIBERO benchmark, we find that policies trained without Cocos tend to omit 3rd-person views, relying almost exclusively on the wrist-mounted 1st-person camera. To quantify this, we compute cosine similarity between randomly sampled visual features extracted before fusion into the policy network. As shown in \Cref{fig:libero_case_study} bottom table, models without Cocos produce highly similar 3rd-person embeddings across diverse scenes, indicating weak discriminability, while Cocos-trained models show significantly greater variation, suggesting stronger visual grounding.

This contrast is clearly illustrated through the rollout comparisons in \Cref{fig:libero_case_study}. (Row 1) shows a successful execution by the model trained with Cocos: the robot pushes the plate forward, turns smoothly to the left, and finally pushes the plate backward to the stove, demonstrating accurate spatial reasoning and multi-step planning. This trajectory serves as a reference for the expected behavior. In (Row 2), a failure case from the model without Cocos, the robot accidentally releases the plate mid-way and attempts to recover by locating it again. However, since the plate is no longer visible in the 1st-person view, the policy fails to re-establish contact, even though the 3rd-person view clearly shows the plate’s new location. This indicates that the model underutilizes the auxiliary camera input. (Row 3) presents a similar failure condition, but the plate remains within the first-person field of view after being released. In this case, the robot successfully relocates and completes the task. Comparing Row 2 and Row 3 highlights the model’s strong reliance on the 1st-person view: if the object leaves that view, recovery becomes unlikely. (Row 4) shows another typical failure from the non-Cocos model: after reaching the stove, the robot continues to push the plate beyond the goal region. This happens because the 1st-person view offers no clear signal indicating proximity to the stove, while the 3rd-person view does. Again, the model fails to leverage this available information. Taken together, these cases reveal a consistent pattern: policies trained without Cocos tend to over-rely on the wrist-mounted camera and struggle to utilize 3rd-person observations. In contrast, models trained with Cocos seldom exhibit these failures, suggesting that the policy network learns to better incorporate multi-view context and is more robust to visual uncertainty.

\begin{figure}[t]
    \centering
    \includegraphics[width=1.0\linewidth]{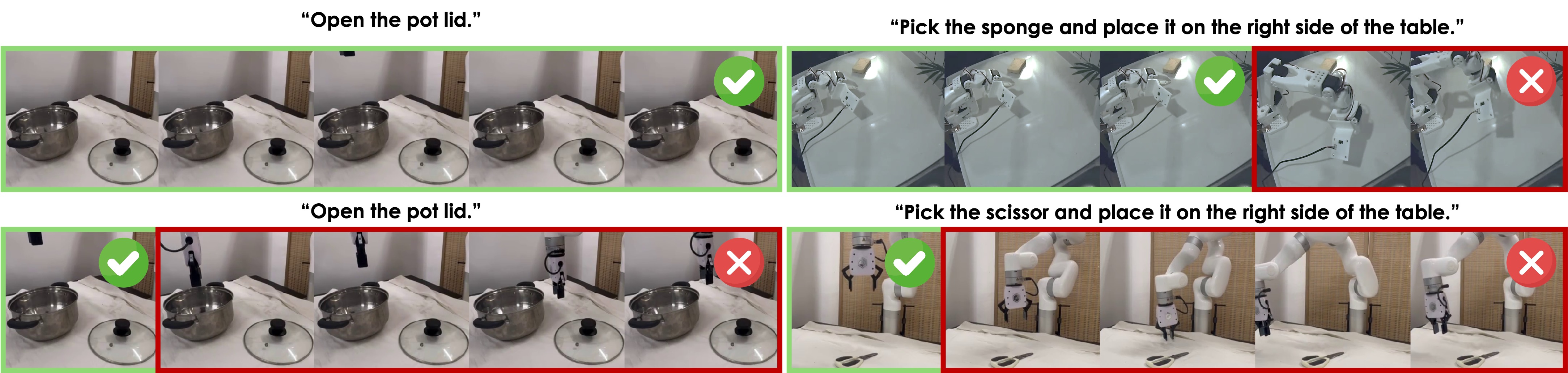}
    \vspace{-15pt}
    \caption{\small{\textbf{Real robot case study.} Top left rollout comes from policy w/ Cocos and the others are from baselines.}}
    \label{fig:case_study_real}
\end{figure}

\textcolor{RedOrange}{\textbf{\textit{Finding 2:}}} \textbf{Cocos reduces modality over-reliance and improves contextual awareness in real-world settings.} In real-robot experiments, we observe that models without Cocos frequently exhibit modality over-reliance, particularly favoring language instructions or arm states over visual cues. As shown in \Cref{fig:case_study_real}, when tasks begin in an already-completed configuration (e.g., lid already open), the baseline policy often executes a full rollout regardless of the visual scene (left bottom). This indicates that the model has learned to follow the instruction without verifying whether it is still applicable. By contrast, policy trained with Cocos correctly remains idle in such states, indicating improved contextual awareness (left top). Similar trends appear in other tasks such as \textit{SpongeMoveRight} (right top) and \textit{ScissorMoveRight} (right bottom), where the baseline struggles to locate target objects or misinterprets spatial references. These errors are less frequent with Cocos, which appears to enhance the policy’s sensitivity to both visual and semantic nuances in the scene. Together, these case studies provide concrete support for our theoretical claims: Cocos prevents loss collapse not just in training metrics, but in how models behave under ambiguous or weakly conditioned scenarios.

\subsection{Ablation Studies}

Cocos only requires the source distribution to be condition-dependent, i.e., $q(z|c)$, and admits flexibility in its concrete form. For simplicity, we instantiate it as a Gaussian with fixed standard deviation and a condition-dependent mean (\Cref{eq:cocos}). Our default setting uses $\beta=0.2$, and we compare it against $\beta=0.1$ and $\beta=0.4$ to investigate the impact of the standard deviation. Inspired by the similarity between our formulation and the variational autoencoder (VAE) posterior, we evaluate a VAE-based version of Cocos, where the standard deviation is automatically learned from data. As shown in \Cref{tab:libero_results}, we observe that the default ($\beta=0.2$) and the wider variant ($\beta=0.4$), and the VAE version all achieve comparable performance across LIBERO task suites. However, the model with $\beta=0.1$ performs substantially worse, suggesting that overly concentrated $q(z|c)$ hinders training. Overall, these results indicate that Cocos is robust to the choice of $\beta$, as long as $q(z|c)$ remains sufficiently spread. 

We also compare the default two-stage training approach, where the source distribution is pre-trained and then fixed, with a joint training strategy using exponential moving average (EMA) updates. In this version, a slowly updated target encoder $F_{\phi^-}$ is used during policy training to provide stable condition-dependent sampling. This online approach achieves performance on par with the two-stage method while simplifying the pipeline, making it a practical choice for integrated training scenarios.

\section{Conclusion, Limitations, and Future Works}
\label{sec:conclusion}
In this paper, we identify and address a critical failure mode in diffusion policy training, loss collapse, where the model fails to distinguish between generative conditions and degenerates into modeling marginal action distributions. We propose Cocos, a simple yet effective modification that injects condition-awareness directly into the source distribution. We provide a theoretical analysis showing how Cocos prevents loss collapse and demonstrate its empirical benefits across extensive simulations and real-world tasks. Our results show that Cocos improves both training efficiency and policy performance, enabling even compact models to match or exceed the performance of large-scale pretrained VLA systems. Despite its effectiveness, Cocos leaves open several important directions. First, while the method only requires the source distribution to be condition-dependent ($q(z|c)$), we have so far instantiated it using a fixed-variance Gaussian derived from autoencoding the condition. Exploring more expressive or learnable source distributions, such as flow-based or attention-guided priors, may further enhance its adaptability. Second, our current experiments are focused on multi-task imitation learning. Extending Cocos to large-scale pretraining and evaluating its ability to generalize across broad VLA domains remains a promising area for future work.

\bibliography{neurips_2025}
\bibliographystyle{plainnat}

\newpage

\appendix

\section{Proofs of Theorems}

\begin{autoproof}{thm:1}
    As in \citet{tong2024improving} we assume that $q, p_t(x|z)$ are decreasing to zero at sufficient speed as $\Vert x\Vert\rightarrow\infty$ and that $u_t, v_t,\nabla_\theta v_t$ are bounded.
    \begin{multline}
        \nabla_\theta\mathbb E_{p_t(x|c)}\Vert v_\theta(t,x,c)-u_t(x|c)\Vert^2=\\\nabla_\theta\mathbb E_{p_t(x|c)}\left(\VVert{v_\theta(t,x,c)}^2-2\AAngle{v_\theta(t, x, c),u_t(x|c)}+\VVert{u_t(x|c)}^2\right)\\
        =\nabla_\theta\mathbb E_{p_t(x|c)}\left(\VVert{v_\theta(t,x,c)}^2-2\AAngle{v_\theta(t, x, c),u_t(x|c)}\right)
    \end{multline}
    \begin{multline}
        \nabla_\theta\mathbb E_{q(z),p_t(x|z,c)}\Vert v_\theta(t,x,c)-u_t(x|z,c)\Vert^2=\\\nabla_\theta\mathbb E_{q(z),p_t(x|z,c)}\left(\VVert{v_\theta(t,x,c)}^2-2\AAngle{v_\theta(t, x, c),u_t(x|z,c)}+\VVert{u_t(x|z,c)}^2\right)\\
        =\nabla_\theta\mathbb E_{q(z),p_t(x|z,c)}\left(\VVert{v_\theta(t,x,c)}^2-2\AAngle{v_\theta(t, x, c),u_t(x|z,c)}\right)
    \end{multline}
    By bilinearity of the 2-norm and since $u_t$ is independent of $\theta$. Next,
    \begin{align}
        \mathbb E_{p_t(x|c)}\VVert{v_\theta(t,x,c)}^2&=\int\VVert{v_\theta(t,x,c)}^2p_t(x|c)\dd x \\
        &=\iint\VVert{v_\theta(t,x,c)}^2p_t(x|z,c)q(z)\dd z\dd x\\
        &=\mathbb E_{q(z),p_t(x|z,c)}\VVert{v_\theta(t,x,c)}^2
    \end{align}
    Finally,
    \begin{align}
        \mathbb E_{p_t(x|c)} \AAngle{v_\theta(t, x, c),u_t(x|c)}&=\int \AAngle{v_\theta(t, x, c),u_t(x|c)}p_t(x|c)\dd x \\
        &=\int \AAngle{v_\theta(t, x, c),\mathbb E_{q(z)}\frac{u_t(x|z,c)p_t( x|z,c)}{p_t(x|c)}}p_t(x|c)\dd x \\
        &=\int \AAngle{v_\theta(t, x, c),\int u_t(x|z,c)p_t( x|z,c)q(z)}\dd z\dd x \\
        &=\iint \AAngle{v_\theta(t, x, c),u_t(x|z,c)}p_t( x|z,c)q(z)\dd z\dd x \\
        &=\mathbb E_{q(z),p_t(x|z,c)}\AAngle{v_\theta(t, x, c),u_t(x|z,c)}
    \end{align}
    Since the gradients of the two objectives are equal for any time $t$, the two objectives are equal.
\end{autoproof}

\begin{autoproof}{thm:2}
  Define
  \begin{equation}
      d(t,x,c)\coloneqq v_\theta(t,x,c)-u_t\bigl(x| x_1,x_0\bigr),
    \quad
    f(z,c)\coloneqq \nabla_\theta v_\theta(t,x,c)^\top\,d(t,x,c),
  \end{equation}
  where we write $y=(x_0,x_1,x)$.  Introduce two measures:
  \begin{itemize}
    \item \emph{Independent measure}
    \begin{equation}
        \mu(\mathrm{d}y)
      =q(x_1,c)\,q(x_0)\,p_t\bigl(x| x_1,x_0\bigr)\,\mathrm{d}x_0\,\mathrm{d}x_1\,\mathrm{d}x,
    \end{equation}
    which does \emph{not} depend on the particular choice of $c$, corresponding to the design choice where $q(z)=q(x_1,c)q(x_0)$.

    \item \emph{Conditional measure}
    \begin{equation}
        \mu_c(\mathrm{d}y)
      =q(x_1,c)\,q\bigl(x_0| c\bigr)\,p_t\bigl(x| x_1,x_0\bigr)\,\mathrm{d}x_0\,\mathrm{d}x_1\,\mathrm{d}x,
    \end{equation}
    which \emph{does} depend on \(c\), corresponding to the design choice where $q(z)=q(x_1,c)q(x_0|c)$.

  \end{itemize}

  We assume the following bounds hold for all \(y\) and \(c\in\mathcal C\):
  \begin{equation}
      \bigl\|\nabla_\theta v_\theta(t,x,c)\bigr\|\le M,
    \quad
    \bigl\|d(t,x,c)\bigr\|\le D,
  \end{equation}
  and that the model’s gradient is \emph{output‐sensitive Lipschitz}, namely there exists \(K>0\) such that
  \begin{equation}
      \bigl\|\nabla_\theta v_\theta(t,x,c_1)
         -\nabla_\theta v_\theta(t,x,c_2)\bigr\|
    \le K\,\bigl\|v_\theta(t,x,c_1)-v_\theta(t,x,c_2)\bigr\|
    \le K\,\epsilon.
  \end{equation}
  
  \medskip\noindent\textbf{(i) Independent measure.}
  Under \(\mu\), the gradient is
  \begin{equation}
      \nabla_\theta\mathcal L_{\mathrm{CFMc}}(\theta,c)
    =2\int f(z,c)\,\mu(\mathrm{d}y).
  \end{equation}
  Hence for any \(c_1,c_2\):
  \begin{align}
    \MoveEqLeft
    \Bigl\|\nabla_\theta\mathcal L_{\mathrm{CFMc}}(\theta,c_1)
    -\nabla_\theta\mathcal L_{\mathrm{CFMc}}(\theta,c_2)\Bigr\| \\
    &=2\Bigl\|\int f(z,c_1)\,\mu(\mathrm{d}y)
      -\int f(z,c_2)\,\mu(\mathrm{d}y)\Bigr\|\\
    &=2\Bigl\|\int\bigl[f(z,c_1)-f(z,c_2)\bigr]\mu(\mathrm{d}y)\Bigr\|.  
  \end{align}
  Expanding the integrand,
  \begin{align}
    f(z,c_1)-f(z,c_2)
    &=\nabla_\theta v_\theta(t,x,c_1)^\top\,d(t,x,c_1)
     -\nabla_\theta v_\theta(t,x,c_2)^\top\,d(t,x,c_2)\\
    &=\nabla_\theta v_\theta(t,x,c_1)^\top\bigl[d(t,x,c_1)-d(t,x,c_2)\bigr] \\
     &+\bigl[\nabla_\theta v_\theta(t,x,c_1) -\nabla_\theta v_\theta(t,x,c_2)\bigr]^\top d(t,x,c_2).
  \end{align}
  Using the bounds \(\|d(c_1)-d(c_2)\|\le\epsilon\) and 
  \(\|\nabla_\theta v(c_1)-\nabla_\theta v(c_2)\|\le K\epsilon\), we obtain
  \begin{align}
    \Bigl\|\nabla_\theta\mathcal L_{\mathrm{CFMc}}(\theta,c_1)
    -\nabla_\theta\mathcal L_{\mathrm{CFMc}}(\theta,c_2)\Bigr\|
    &\le2\int\Bigl(
      M\,\epsilon + K\,\epsilon\,D
    \Bigr)\,\mu(\mathrm{d}y)\\
    &=2\,(M+K D)\,\epsilon.
  \end{align}

    The above bound shows that when all $c \in \mathcal C$ induce similar gradients, the network update over this region becomes nearly invariant to $c$. In the limit of gradient descent, the parameter $\theta$ thus evolves according to a shared direction, regardless of the exact $c$ value. As a result, the learned model $v_\theta(t,x,c)$ tends to converge, for all $c\in\mathcal C$, to a $c$-independent function $v^*(t,x)$ that minimizes the averaged objective:
    \begin{equation}
        v^*(t,x) := \arg\min_{v}~\mathbb{E}_{c\in\mathcal C} \mathbb{E}_{z\sim\mu} \left[\|v(t,x) - u_t(x|x_1,x_0)\|^2\right].
    \end{equation}
    That is, the network effectively collapses the conditional dependency on $c$ and behaves like a shared estimator $v^*$ optimized for average performance across $\mathcal C$.
    
  \medskip\noindent\textbf{(ii) Conditional measure.}
  Under \(\mu_c\), the gradient becomes
  \begin{equation}
      \nabla_\theta\mathcal L_{\mathrm{CFMc}}(\theta,c)
    =2\int f(z,c)\,\mu_c(\mathrm{d}y).
  \end{equation}
  For \(c_1,c_2\), write
  \begin{align}
    &\nabla_\theta\mathcal L_{\mathrm{CFMc}}(\theta,c_1)
     -\nabla_\theta\mathcal L_{\mathrm{CFMc}}(\theta,c_2)\\
    &=2\Bigl[\int f(z,c_1)\,\mu_{c_1}(\mathrm{d}z)
      -\int f(z,c_2)\,\mu_{c_2}(\mathrm{d}z)\Bigr]\\
    &=2\Bigl[\bigl(\int f(z,c_1)\,\mu_{c_1}(\mathrm{d}z)
      -\int f(z,c_1)\,\mu_{c_2}(\mathrm{d}z)\bigr)
      +\bigl(\int f(z,c_1)\,\mu_{c_2}(\mathrm{d}z)
      -\int f(z,c_2)\,\mu_{c_2}(\mathrm{d}z)\bigr)\Bigr]\\
    &\;=\;T_1 + T_2.
  \end{align}
  The second term \(T_2\) coincides with the independent‐measure analysis and thus is bounded by \(2(M+KD)\,\epsilon\). The first term
  \begin{equation}
      T_1
      =2\Bigl(\int f(z,c_1)\,\mu_{c_1}(\mathrm{d}z)
      -\int f(z,c_1)\,\mu_{c_2}(\mathrm{d}z)\Bigr)
  \end{equation}
is the difference of the same function under two \emph{different} measures.  Without any further assumption on the relationship between \(\mu_{c_1}\) and \(\mu_{c_2}\), this difference can be made arbitrarily large.  For instance, if \(q(x_0| c_1)=\delta_a\) and \(q(x_0| c_2)=\delta_b\), then \(\|T_1\|=2|f(a,c_1)-f(b,c_1)|\) can diverge independently of \(\epsilon\). Hence, under the conditional measure \(\mu_c\), no bound depending only on \(\epsilon\) exists for \(\|\nabla_\theta L(\theta,c_1)-\nabla_\theta L(\theta,c_2)\|\).
\end{autoproof}

\section{Details of Benchmarks}\label{appendix:benchmarks}

\begin{figure}[ht]
    \centering
    \includegraphics[width=1.0\linewidth]{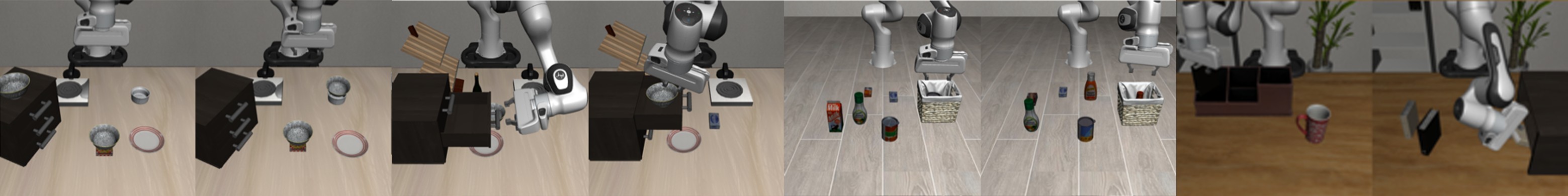}
    \vspace{-10pt}
    \caption{\small \textbf{LIBERO simulation benchmark.} We conduct experiments on 40 tasks from four task suites in the LIBERO benchmark. We show two task examples for each suite here.}
    \label{fig:libero_task}
\end{figure}

\begin{figure}[ht]
    \centering
    \includegraphics[width=1.0\linewidth]{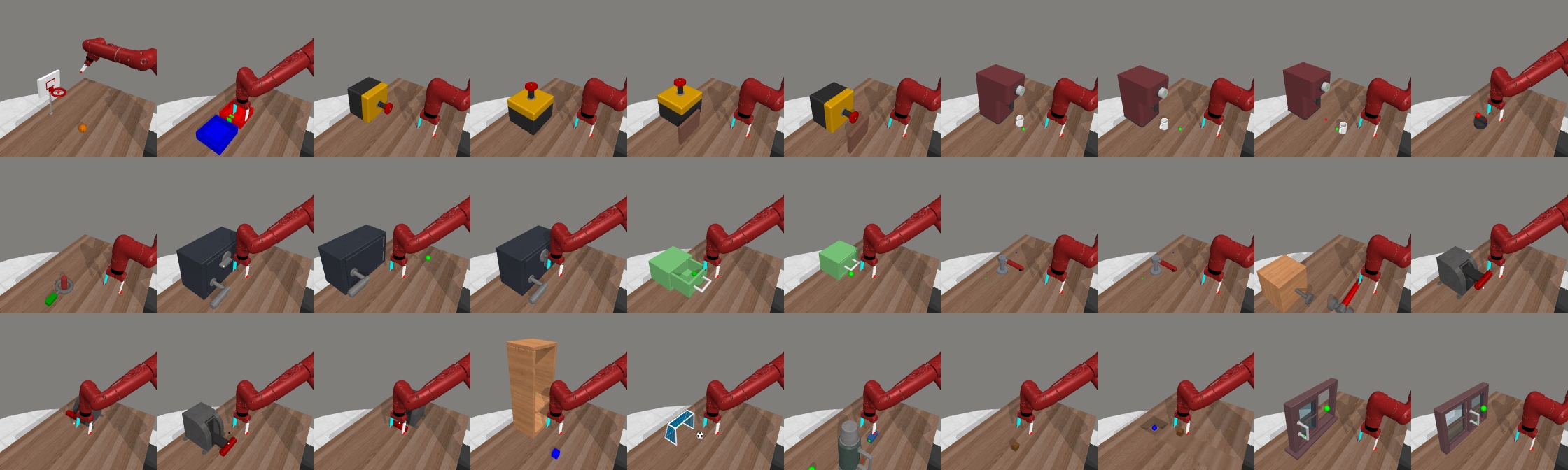}
    \caption{\small \textbf{MetaWorld simulation benchmark.} We conduct experiments on 30 tasks of three difficulty levels in the MetaWorld benchmark. We show all task examples here.}
    \label{fig:metaworld_task}
\end{figure}

\begin{figure}[ht]
    \centering
    \includegraphics[width=1.0\linewidth]{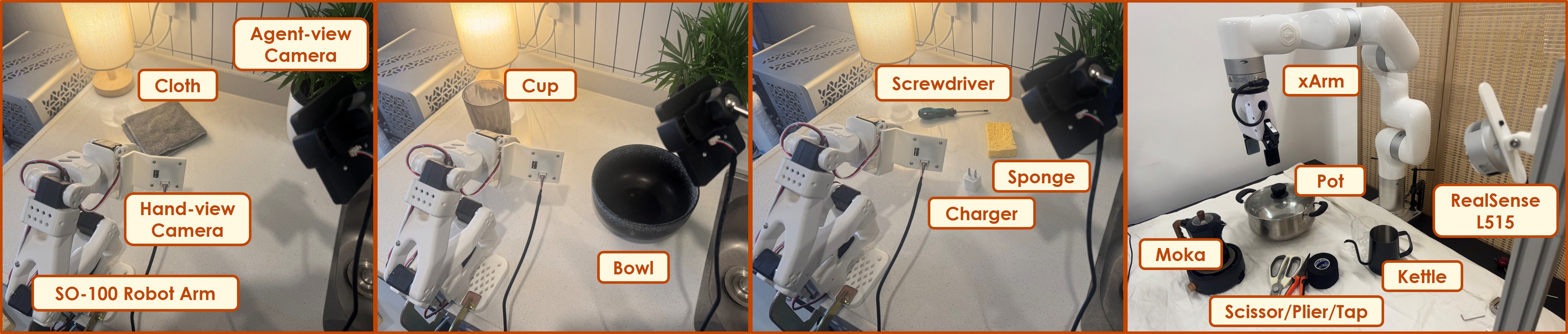}
    \vspace{-10pt}
    \caption{\small{\textbf{Real-world experimental setups.} We conduct experiments on both SO100 and xArm platform. For each robot, we design a suite of 10 tabletop tasks involving diverse objects.}}
    \label{fig:so100_task}
\end{figure}

\textbf{LIBERO.} The LIBERO simulation benchmark \citep{liu2023libero} features a Franka Emika Panda arm in simulation across four challenging task suites: \textit{Goal}, \textit{Spatial}, \textit{Object}, and \textit{Long}. Each suite comprises 10 tasks with 500 demonstrations and is designed to investigate controlled knowledge transfer related to goal variations, spatial configurations, object types, and long-horizon tasks. Unlike prior work \citep{kim2024openvla, kim2025openvlaoft}, we do not filter out unsuccessful demonstrations, aiming for a more realistic evaluation setting. For policy training, the model predicts action chunks of length 16; after each chunk prediction, 8 steps are executed before generating the next chunk. The observation space includes 2-view RGB images at the current time step, without historical observations. During evaluation, following \citet{liu2023libero}, each task is tested over 50 trials with 3 different random seeds, and success rates are reported. To provide a clearer understanding of the task suites, we present agent-view observations in \Cref{fig:libero_task} and detailed task descriptions in \Cref{tab:libero_task}.

\textbf{MetaWorld.} The MetaWorld simulation benchmark \citep{yu2019metaworld} includes 50 distinct tabletop manipulation tasks using a Sawyer robot arm. We select 30 tasks from \textit{easy}, \textit{medium}, and \textit{very hard} difficulty levels to evaluate VLA models. We use a scripted policy to collect 20 demonstrations for each task. For policy training, the model predicts action chunks of length 16; after each chunk prediction, 16 steps are executed before generating the next chunk. The observation space consists of a single RGB image at the current time step, without historical observations. During evaluation, each task is tested over 50 trials with 3 different random seeds, and success rates are reported. To better illustrate the task suites, we show agent-view observations in \Cref{fig:metaworld_task} and task descriptions in \Cref{tab:metaworld_task}.

\textbf{SO100 Robot Manipulation.} The SO100 robot \citep{cadene2024lerobot} is a low-cost, open-source 6-DoF manipulator, with both the leader and follower arms costing approximately \$250. We assemble the hardware using a 3D-printed kit provided by the open-source community. The robot has two RGB cameras: one mounted on the wrist and the other positioned to provide a third-person view. Both cameras operate at a resolution of 640×480 and 25 FPS. The robot controller runs at 30Hz, and actions are defined as target absolute joint angles. Due to its low-cost design, the platform has several hardware limitations, including significant arm jitter, low load capacity, and occasional camera lag, which present practical challenges for developing embodied AI systems. However, given the increasing adoption of such affordable open-source robots by the research community, we believe that evaluating models on these lower-performance systems offers valuable insights and broader applicability. We design four categories of tabletop manipulation tasks for the SO100 setup: \textit{\textbf{(1) Pick\&Place:}} involving 3 objects and 2 placement zones (6 tasks), \textit{\textbf{(2) MoveTo:}} navigating 2 objects to a single target zone (2 tasks), \textit{\textbf{(3) Wipe:}} picking up a cloth and wiping the table (1 task), and \textit{\textbf{(4) Unfold:}} unfolding a cloth (1 task). In total, we evaluate performance on 10 distinct tasks. Language instructions for each task are listed in \Cref{tab:so100_task}, and visual examples of the task environments are shown in \Cref{fig:so100_task}.

During data collection, we record 20 demonstrations per task. For policy training, the model predicts an action chunk of length 64; after each chunk prediction, 40 steps are executed before generating the next chunk. The observation space includes 2-view RGB images at the current time step, along with the absolute joint angles from the current and previous 10 steps. During evaluation, each task is tested over 10 trials, and success rates are reported.

\textbf{xArm Robot Manipulation.} xArm is a high-performance 7-DoF manipulator. The robot is equipped with a third-person view Intel RealSense L515 LiDAR camera, operating at 640$\times$480 resolution and 30 FPS. We collect both RGB and depth images from the camera. The robot controller runs at 30Hz, and actions are defined as target absolute joint angles. We design four categories of tabletop manipulation tasks for the xArm setup: \textit{\textbf{(1) Pick\&Place:}} involving 3 objects and 2 placement zones (6 tasks), \textit{\textbf{(2) Pot:}} taking off or putting on the pot lid (2 tasks), \textit{\textbf{(3) Pour:}} pouring water from the kettle into the cup (1 task), and \textit{\textbf{(4) Moka:}} placing the Moka pot on the cooker (1 task). In total, we evaluate performance on 10 distinct tasks. Language instructions for each task are listed in \Cref{tab:so100_task}, and visual examples of task environments are shown in \Cref{fig:so100_task}.

During data collection, we record 20 demonstrations per task. For policy training, the model predicts an action chunk of length 64; after each chunk prediction, 40 steps are executed before generating the next chunk. The observation space includes a third-person view RGB image at the current time step, as well as the absolute joint angles from the current and previous 10 steps. During evaluation, each task is tested over 10 trials, and success rates are reported.

\begin{table}[htbp]
\centering
\caption{{\small Task description of each task in the LIBERO benchmark.}}
\scalebox{0.8}{
\label{tab:libero_task}
\begin{tabular}{l|l}
\toprule
\multicolumn{1}{c|}{Task Suite}  & \multicolumn{1}{c}{Task Description}                                                       \\
\midrule
\multirow{10}{*}{LIBERO-Goal}    & \small open the middle layer of the drawer                                                        \\
                                 & \small put the bowl on the stove                                                                  \\
                                 & \small put the wine bottle on the top of the drawer                                               \\
                                 & \small open the top layer of the drawer and put the bowl inside                                   \\
                                 & \small put the bowl on the top of the drawer                                                      \\
                                 & \small push the plate to the front of the stove                                                   \\
                                 & \small put the cream cheese on the bowl                                                           \\
                                 & \small turn on the stove                                                                          \\
                                 & \small put the bowl on the plate                                                                  \\
                                 & \small put the wine bottle on the rack                                                            \\
\midrule
\multirow{10}{*}{LIBERO-Spatial} & \small pick the akita black bowl between the plate and the ramekin and place it on the plate      \\
                                 & \small pick the akita black bowl next to the ramekin and place it on the plate                    \\
                                 & \small pick the akita black bowl from table center and place it on the plate                      \\
                                 & \small pick the akita black bowl on the cookies box and place it on the plate                     \\
                                 & \small pick the akita black bowl in the top layer of the wooden cabinet and place it on the plate \\
                                 & \small pick the akita black bowl on the ramekin and place it on the plate                         \\
                                 & \small pick the akita black bowl next to the cookies box and place it on the plate                \\
                                 & \small pick the akita black bowl on the stove and place it on the plate                           \\
                                 & \small pick the akita black bowl next to the plate and place it on the plate                      \\
                                 & \small pick the akita black bowl on the wooden cabinet and place it on the plate                  \\
\midrule
\multirow{10}{*}{LIBERO-Object}  & \small pick the alphabet soup and place it in the basket                                          \\
                                 & \small pick the cream cheese and place it in the basket                                           \\
                                 & \small pick the salad dressing and place it in the basket                                         \\
                                 & \small pick the bbq sauce and place it in the basket                                              \\
                                 & \small pick the ketchup and place it in the basket                                                \\
                                 & \small pick the tomato sauce and place it in the basket                                           \\
                                 & \small pick the butter and place it in the basket                                                 \\
                                 & \small pick the milk and place it in the basket                                                   \\
                                 & \small pick the chocolate pudding and place it in the basket                                      \\
                                 & \small pick the orange juice and place it in the basket                                           \\
\midrule
\multirow{10}{*}{LIBERO-Long}    & \small put both the alphabet soup and the tomato sauce in the basket                              \\
                                 & \small put both the cream cheese box and the butter in the basket                                 \\
                                 & \small turn on the stove and put the moka pot on it                                               \\
                                 & \small put the black bowl in the bottom drawer of the cabinet and close it                        \\
                                 & \small put the white mug on the left plate and put the yellow and white mug on the right plate    \\
                                 & \small pick up the book and place it in the back compartment of the caddy                         \\
                                 & \small put the white mug on the plate and put the chocolate pudding to the right of the plate     \\
                                 & \small put both the alphabet soup and the cream cheese box in the basket                          \\
                                 & \small put both moka pots on the stove                                                            \\
                                 & \small put the yellow and white mug in the microwave and close it                                 \\
\bottomrule
\end{tabular}}
\end{table}

\begin{table}[htbp]
  \centering
  \caption{\small Task description of each task in the MetaWorld benchmark.}
  \label{tab:metaworld_task}
  \scalebox{0.8}{
  \begin{tabular}{l|l}
    \toprule
    \textbf{Task Name} & \textbf{Task Description} \\
    \midrule
    basketball & Dunk the basketball into the basket. \\
    bin-picking & Grasp the puck from one bin and place it into another bin. \\
    button-press & Press a button. \\
    button-press-topdown & Press a button from the top. \\
    button-press-topdown-wall & Bypass a wall and press a button from the top. \\
    button-press-wall & Bypass a wall and press a button. \\
    coffee-button & Push a button on the coffee machine. \\
    coffee-pull & Pull a mug from a coffee machine. \\
    coffee-push & Push a mug under a coffee machine. \\
    dial-turn & Rotate a dial 180 degrees. \\
    disassemble & Pick a nut out of the peg. \\
    door-lock & Lock the door by rotating the lock clockwise. \\
    door-open & Open a door with a revolving joint. \\
    door-unlock & Unlock the door by rotating the lock counter-clockwise. \\
    drawer-close & Push and close a drawer. \\
    drawer-open & Open a drawer. \\
    faucet-close & Rotate the faucet clockwise. \\
    faucet-open & Rotate the faucet counter-clockwise. \\
    hammer & Hammer a screw on the wall. \\
    handle-press & Press a handle down. \\
    handle-press-side & Press a handle down sideways. \\
    handle-pull & Pull a handle up. \\
    handle-pull-side & Pull a handle up sideways. \\
    shelf-place & Pick and place a puck onto a shelf. \\
    soccer & Kick a soccer into the goal. \\
    stick-push & Grasp a stick and push a box using the stick. \\
    sweep & Sweep a puck off the table. \\
    sweep-into & Sweep a puck into a hole. \\
    window-close & Push and close a window. \\
    window-open & Push and open a window. \\
    \bottomrule
  \end{tabular}}
\end{table}

\begin{table}[htbp]
\centering
\caption{\small\textbf{Task description of each task in the SO100 and xArm benchmark.} As each parameter combination introduces one task, each task suite has 10 tasks in total. For each task, we test the model for 10 trials.}
\label{tab:so100_task}
\scalebox{0.7}{
\begin{tabular}{l|l|l}
\toprule
Task Suite & Task Description & Parameter \\
\midrule

\multirow{4}{*}{SO100} & 
\small pick \texttt{[A]} and place it on the \texttt{[B]} side of the table & 
\small \texttt{[A]: ["screwdriver", "sponge", "charger"]}, \texttt{[B]: ["left", "right"]} \\
 & 
\small move \texttt{[A]} to the center of the table &
\small \texttt{[A]: ["cup", "bowl"]} \\
& 
\small pick the cloth and wipe the table & 
\small \texttt{None} \\
& 
\small unfold the cloth & 
\small \texttt{None} \\

\midrule

\multirow{4}{*}{xArm} &

\small pick \texttt{[A]} and place it on the \texttt{[B]} side of the table & 
\small \texttt{[A]: ["scissor", "plier", "tap"]}, \texttt{[B]: ["left", "right"]} \\
 & 
\small open the pot lid \texttt{or} put the lid on the pot &
\small \texttt{[open, close]} \\
& 
\small pour the water from the kettle into the cup & 
\small \texttt{None} \\
& 
\small place the Moka pot on the cooker & 
\small \texttt{None} \\

\bottomrule
\end{tabular}}
\end{table}

\section{Details of Evaluation Metrics}
\label{appendix:metrics}

\textbf{Cosine similarity.} To quantify the influence of condition inputs on our policy network, we measure the cosine similarity between hidden states before and after condition injection. Given an $L$-layer RDT policy network where conditions are integrated via cross-attention in each transformer block, we denote the hidden states before and after condition injection as $h^l, \bar{h}^l \in \mathbb{R}^{S \times D}$, where $S$ is the sequence length and $D$ is the transformer hidden dimension. Since earlier layers' transformations propagate through the network, affecting all subsequent computations, we calculate an exponentially weighted sum: $\sum_{l=0}^{L-1} w^l \min_S\{\text{Sim}(h^l, \bar{h}^l)\}$, where $\min_S$ operates across the sequence dimension, $\text{Sim}$ computes cosine similarity across the hidden dimension, and $w=0.5$ gives greater weight to earlier layers. Lower similarity values indicate stronger condition influence on network representations.

\textbf{Norm scale change.} Similarly, we quantify the magnitude of change induced by conditional inputs using an exponentially weighted sum of relative norm changes: $\sum_{l=0}^{L-1} w^l \max_S \left|\frac{\|\bar{h}^l\| - \|h^l\|}{\|h^l\|}\right|$, where $\max_S$ operates across the sequence dimension, $\|\cdot\|$ computes the norm across the hidden dimension, and $w=0.5$. Larger values indicate that conditional information substantially alters the magnitude of hidden representations, suggesting stronger condition integration within the policy network.

\textbf{Cosine similarity in case study.} In \Cref{fig:libero_case_study}, we also calculate a cosine similarity metric to quantify the degree to which the policy network differentiates between conditions. This cosine similarity is computed between visual features \textit{before} they are fused into the policy network (i.e., after the vision encoder's output has been processed by positional encoding and an MLP, yielding the features ultimately used for fusion with the policy network). We calculate this metric by randomly sampling image pairs from the dataset and then averaging the cosine similarity of their corresponding features.

\section{Policy Backbone Recipe}

\subsection{Condition Network}

For vision encoding, we utilize the pre-trained DINOv2-Base model, and for language encoding, we employ the pre-trained T5-Base model. Both of these encoders are frozen during training. The outputs from these pre-trained modality encoders are processed by a linear projector and then augmented with positional encodings before being fused into the policy network. These positional encodings are trainable but are initialized with sin-cos positional encoding and scaled down by a factor of 0.2.

\subsection{Policy Network}

We employ a compact Robot Diffusion Transformer (RDT) architecture, which is fundamentally a Diffusion Transformer incorporating cross-attention layers. Diffusion timestamps and robot kinematic information are integrated into the policy network using AdaLN-Zero \citep{Peebles2022DiT}. The vision and language embeddings are used as the Keys and Values in the cross-attention layers to be integrated into the policy network alternately \citep{liu2025rdt}. The Transformer architecture has a hidden dimension of 384, with 6 attention heads, and 12 layers.

\subsection{Diffusion Model}

We use a flow matching model defined by \Cref{eq:diffusion_implement}. Diffusion timestamps are treated as continuous values within the range $[0, 1]$; we do not discretize them. Instead, they are represented using a Fourier embedding with a scale of 0.2 \citep{dong2024cleandiffuser}. During training, diffusion timestamps are sampled from a uniform distribution over the interval $[0, 1]$. For inference, we solve the corresponding ODE using the Euler method, dividing the interval $[0, 1]$ into equal-sized steps.

\subsection{Computation Resources}

All policy training and testing are conducted on a server equipped with 4 NVIDIA GeForce RTX 4090 GPUs and an Intel(R) Xeon(R) CPU E5-2678 v3 @ 2.50GHz.


\newpage

\section*{NeurIPS Paper Checklist}

\begin{enumerate}

\item {\bf Claims}
    \item[] Question: Do the main claims made in the abstract and introduction accurately reflect the paper's contributions and scope?
    \item[] Answer: \answerYes{} 
    \item[] Justification: The main claims made in the abstract and introduction accurately reflect the paper's contributions and scope.

\item {\bf Limitations}
    \item[] Question: Does the paper discuss the limitations of the work performed by the authors?
    \item[] Answer: \answerYes{} 
    \item[] Justification: We have discussed the limitations of the work in \Cref{sec:conclusion}.

\item {\bf Theory assumptions and proofs}
    \item[] Question: For each theoretical result, does the paper provide the full set of assumptions and a complete (and correct) proof?
    \item[] Answer: \answerYes{} 
    \item[] Justification: We provide the full set of assumptions and a complete proof for each theoretical result.

    \item {\bf Experimental result reproducibility}
    \item[] Question: Does the paper fully disclose all the information needed to reproduce the main experimental results of the paper to the extent that it affects the main claims and/or conclusions of the paper (regardless of whether the code and data are provided or not)?
    \item[] Answer: \answerYes{} 
    \item[] Justification: We provide every detail to reproduce the main experimental results.

\item {\bf Open access to data and code}
    \item[] Question: Does the paper provide open access to the data and code, with sufficient instructions to faithfully reproduce the main experimental results, as described in supplemental material?
    \item[] Answer: \answerYes{} 
    \item[] Justification: We will open-source the code in a few days after submission.

\item {\bf Experimental setting/details}
    \item[] Question: Does the paper specify all the training and test details (e.g., data splits, hyperparameters, how they were chosen, type of optimizer, etc.) necessary to understand the results?
    \item[] Answer: \answerYes{} 
    \item[] Justification: We provide every detail in training and testing.

\item {\bf Experiment statistical significance}
    \item[] Question: Does the paper report error bars suitably and correctly defined or other appropriate information about the statistical significance of the experiments?
    \item[] Answer: \answerYes{} 
    \item[] Justification: Yes. We report error bars in the learning curve.

\item {\bf Experiments compute resources}
    \item[] Question: For each experiment, does the paper provide sufficient information on the computer resources (type of compute workers, memory, time of execution) needed to reproduce the experiments?
    \item[] Answer: \answerYes{} 
    \item[] Justification: Yes. We list the GPU and CPU resources.
    
\item {\bf Code of ethics}
    \item[] Question: Does the research conducted in the paper conform, in every respect, with the NeurIPS Code of Ethics \url{https://neurips.cc/public/EthicsGuidelines}?
    \item[] Answer: \answerYes{} 
    \item[] Justification: Yes, we do.

\item {\bf Broader impacts}
    \item[] Question: Does the paper discuss both potential positive societal impacts and negative societal impacts of the work performed?
    \item[] Answer: \answerYes{} 
    \item[] Justification: Yes, we do.
    
\item {\bf Safeguards}
    \item[] Question: Does the paper describe safeguards that have been put in place for responsible release of data or models that have a high risk for misuse (e.g., pretrained language models, image generators, or scraped datasets)?
    \item[] Answer: \answerNA{} 
    \item[] Justification: The paper poses no such risks.

\item {\bf Licenses for existing assets}
    \item[] Question: Are the creators or original owners of assets (e.g., code, data, models), used in the paper, properly credited and are the license and terms of use explicitly mentioned and properly respected?
    \item[] Answer: \answerYes{} 
    \item[] Justification: Yes, they are.

\item {\bf New assets}
    \item[] Question: Are new assets introduced in the paper well documented and is the documentation provided alongside the assets?
    \item[] Answer: \answerYes{} 
    \item[] Justification: Yes.

\item {\bf Crowdsourcing and research with human subjects}
    \item[] Question: For crowdsourcing experiments and research with human subjects, does the paper include the full text of instructions given to participants and screenshots, if applicable, as well as details about compensation (if any)? 
    \item[] Answer: \answerNA{} 
    \item[] Justification: We do not include crowdsourcing experiments or research with human subjects.

\item {\bf Institutional review board (IRB) approvals or equivalent for research with human subjects}
    \item[] Question: Does the paper describe potential risks incurred by study participants, whether such risks were disclosed to the subjects, and whether Institutional Review Board (IRB) approvals (or an equivalent approval/review based on the requirements of your country or institution) were obtained?
    \item[] Answer: \answerNA{} 
    \item[] Justification: We do not include crowdsourcing experiments or research with human subjects.

\item {\bf Declaration of LLM usage}
    \item[] Question: Does the paper describe the usage of LLMs if it is an important, original, or non-standard component of the core methods in this research? Note that if the LLM is used only for writing, editing, or formatting purposes and does not impact the core methodology, scientific rigorousness, or originality of the research, declaration is not required.
    \item[] Answer: \answerNA{} 
    \item[] Justification: The core method development in this research does not involve LLMs.

\end{enumerate}

\end{document}